\documentclass[conference]{IEEEtran}
\IEEEoverridecommandlockouts
\usepackage{cite}
\usepackage{lipsum}
\usepackage{amsmath,amssymb,amsfonts}
\usepackage{algorithmic}
\usepackage{booktabs}
\usepackage{graphicx}
\usepackage{textcomp}
\usepackage{xcolor}
\usepackage{threeparttable}
\def\BibTeX{{\rm B\kern-.05em{\sc i\kern-.025em b}\kern-.08em
    T\kern-.1667em\lower.7ex\hbox{E}\kern-.125emX}}
\begin{document}


\title{Seeing Like a Designer Without One: A Study on Unsupervised Slide Quality Assessment via Designer Cue Augmentation\\
}

\author{\IEEEauthorblockN{Tai Inui}
\IEEEauthorblockA{Waseda University \\
Tokyo, Japan \\
taiinui556@suou.waseda.jp}

\and

\IEEEauthorblockN{Steven Oh}
\IEEEauthorblockA{Waseda University \\
Tokyo, Japan \\
oh.steven@fuji.waseda.jp}

\and

\IEEEauthorblockN{Magdeline Kuan}
\IEEEauthorblockA{Waseda University \\
Tokyo, Japan \\
magdeline.kuan@akane.waseda.jp}
}

\maketitle

\begin{abstract}

We present an unsupervised slide‐quality assessment pipeline that combines seven expert‐inspired visual‐design metrics (whitespace, colorfulness, edge density, brightness contrast, text density, color harmony, layout balance) with CLIP‑ViT embeddings, using Isolation Forest–based anomaly scoring to evaluted presentation slides. Trained on 12k professional lecture slides and evaluated on six academic talks (115 slides), our method achieved Pearson correlations up to 0.83 with human visual‐quality ratings—1.79× to 3.23× stronger than scores from leading vision–language models (ChatGPT o4‑mini‑high, ChatGPT o3, Claude Sonnet 4, Gemini 2.5 Pro). We demonstrate convergent validity with visual ratings, discriminant validity against speaker‐delivery scores, and exploratory alignment with overall impressions. Our results show that augmenting low‐level design cues with multimodal embeddings closely approximates audience perceptions of slide quality, enabling scalable, objective feedback in real time. 
\end{abstract}

\begin{IEEEkeywords}
presentation, design, slides, visual, assessment
\end{IEEEkeywords}

\section{Introduction}
Slideware such as PowerPoint, Keynote and Google Slides has become the primary visual channel in classrooms, boardrooms and pitch competitions. Audience‐tracking studies show that, in a typical talk, viewers spend a majority of their gaze time looking at the projected slides rather than the speaker, making slide quality a dominant driver of comprehension and persuasion \cite{structureness2023}. Experimental evidence supports this as engineering students who viewed assertion-evidence slides that followed multimedia-learning principles scored higher on immediate and delayed recall tests and reported lower cognitive load than peers who saw conventional “bullet-point” decks \cite{garner2012assertion}. 

In business settings, investors also rely heavily on visual cues: something as specific as the clarity of a growth-timeline graphic can change funding decisions in early-stage pitch decks \cite{blaseg2025visualpitch}. Taken together, these findings imply that any holistic evaluation of presentation skill must include an objective appraisal of the slide deck itself. Yet, current feedback practices remain largely subjective and labor-intensive, leaving presenters with little real-time guidance and researchers with noisy evaluation baselines. This paper addresses that gap by proposing a machine learning pipeline that rates slides along objective design dimensions based on expert design guidelines: whitespace, color richness, edge density, brightness contrast, text density, color harmony and layout balance, in addition to comparing the model’s evaluation to presentation evaluation survey responses.

\begin{figure*}
    \centering
    \includegraphics[width=0.8\linewidth]{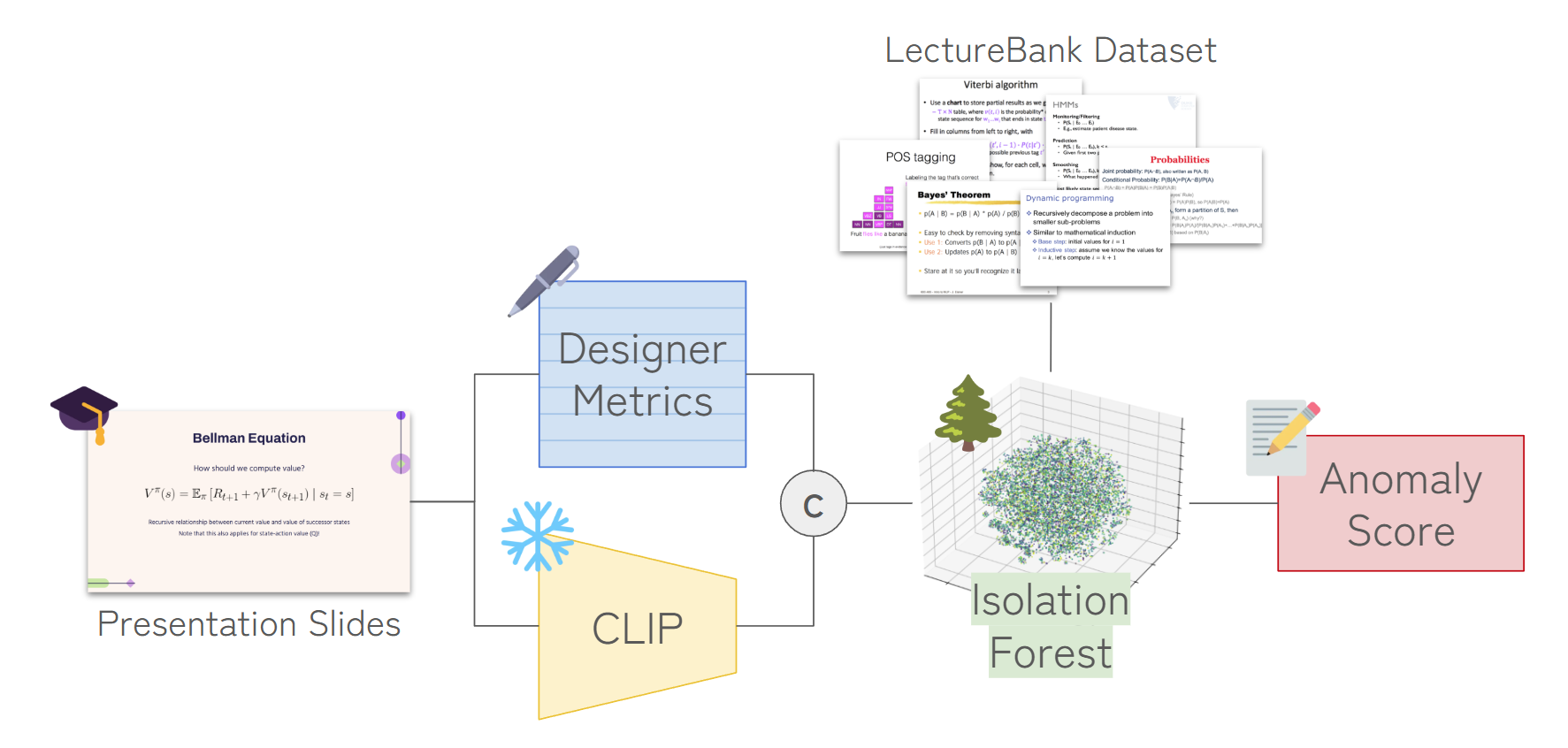}
    \caption{\textbf{Overview of our approach.} We take seven design metrics and CLIP embeddings from a presentation slide, then concatenate them to create a latent space that represents both high and low-level visual features of a slide. We then compare it with the professional slide distribution from the LectureBank dataset via isolation forest to obtain an anomaly score.}
    \label{fig:pipeline}
\end{figure*}

\section{Related Work}

\subsection{Slides as a Determinant of Presentation Effectiveness}

Cognitive-theory work dating back to Mayer shows that well-designed multimedia displays off-load working memory by aligning words and images, whereas cluttered screens impose extraneous load that suppresses learning. Garner \& Alley’s classroom study operationalized those principles for slides and demonstrated measurable learning gains when they were respected \cite{garner2012assertion}. Beyond education, audience-engagement research with undergraduate listeners found that the structuredness of slide content, including clear hierarchy and consistent visual rhythm, significantly boosts both engagement ratings and mastery of the material \cite{structureness2023}. Usability scholars reach similar conclusions from a human–computer-interaction angle: Nielsen’s heuristic \#8 (“aesthetic and minimalist design”) links sparse, well-balanced visuals to higher task efficiency and lower error rates, underscoring whitespace and visual economy as design levers that matter just as much on a slide as on a screen interface \cite{nielsen1994heuristics}.

\subsection{Visual Design Variables Relevant to Automated Scoring}

Design literature highlights whitespace as essential for guiding visual flow and enhancing professionalism; generous margins improve readability and perceived clarity\cite{mayer2005cognitive}. Text density is similarly important—concise slides with limited text aid comprehension and reduce cognitive overload\cite{naegle2021slides}. Our system quantifies both to reward clarity and simplicity.

For color, we include metrics for harmony and colorfulness: perceptual studies show that balanced hues and moderate vibrancy increase aesthetic appeal and attention without overwhelming viewers\cite{zhang2022color}\cite{zhou2022colorharmony}. Edge density acts as a proxy for visual clutter, where lower complexity supports better focus, while brightness contrast improves text legibility and content emphasis\cite{rosenholtz2007measuring}.

Lastly, we assess layout balance as studies in interface design show that symmetrical and evenly distributed visual elements improve perceived aesthetics and ease of processing. Balanced layouts help guide the viewer’s eye, reduce cognitive load, and enhance the professional appearance of a slide\cite{ngo2003modelling}.

\subsection{Autonomous Slide Quality Evaluation}

Early automatic tools relied on handcrafted rules or style guides, which struggled with diverse templates and trade-offs between competing aesthetics. Data-driven efforts began with Kim et al.’s Quality-Based Automatic Classification for Presentation Slides, which framed information-quality detection as a supervised classification task and achieved promising accuracy with global slide features \cite{kim2014classification}. Hanani et al. extended the idea to multimodal analytics, showing that models trained solely on slide attributes—word count, graphic density, maximum font size—could predict instructors’ presentation grades with up to 65\% accuracy and that fusing slide, audio and gesture features further improved performance \cite{multimodal2017}. 

More recent efforts like LecEval \cite{leceval2024} and PPTEval / PPTAgent \cite{pptagent2024} take a different approach, relying on subjective human or LLM-based evaluation. LecEval \cite{leceval2024} incorporates gaze, voice, and gesture data to estimate how effective a slide is in educational contexts, but focuses more on high-level lecture delivery than on visual design principles. PPTEval \cite{pptagent2024} uses GPT-4o to rate slide quality based on criteria like coherence, informativeness, and conciseness; however, it does not explicitly assess concrete visual metrics such as alignment, whitespace distribution, or color harmony. Both systems offer holistic judgments, but lack objective, design-focused feedback grounded in measurable layout and stylistic attributes.

\section{Method}

Our pipeline combines low‑level slide design heuristics with high‑level vision–language representations to produce a single, unsupervised quality score. Specifically, we extract seven interpretable design metrics from each slide image, embed the same image with CLIP, fuse the two representations through latent‑space augmentation, and finally compute an anomaly score with isolation forest on a corpus of professional slides (Figure \ref{fig:pipeline}).

\subsection{Dataset}
We sampled slide PDFs from the LectureBank dataset \cite{li2018lecturebank}, which contains lecture slides from 5 different academic domains: Natural Language Processing, Machine Learning, Artificial Intelligence, Deep Learning, and Information Retrieval. For our work, this set is treated as examples of “expert” academic design to compare against. We rasterized the PDFs into images, resulting in 12k total images.

\subsection{Design Metrics}
We implement seven metrics motivated by graphic‑design literature: Whitespace, Text Density, Color Harmony, Colorfulness, Edge Density, Brightness Contrast, and Layout Balance. Each metric is calculated with lightweight image processing operators (for example, pixel intensity thresholds, Canny edges) and normalized so that higher numbers always indicate more of the property, but are not themselves 'good' or 'bad'.
\subsection{High‑level visual encoding}

\begin{figure}
    \centering
    \includegraphics[width=1\linewidth]{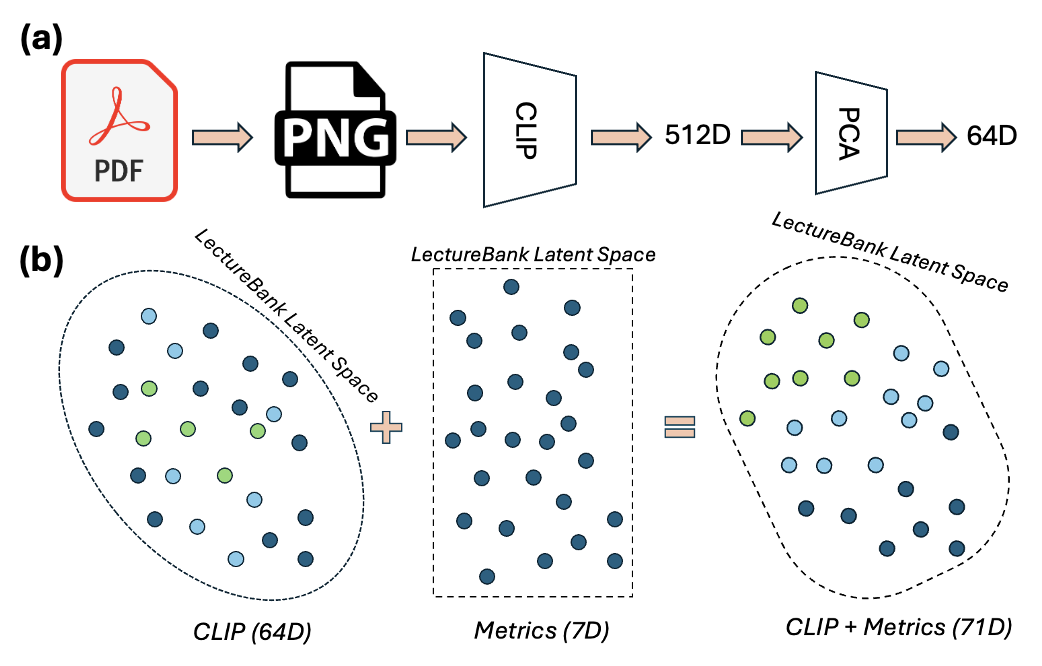}
    \caption{(a) LectureBank slide‐data processing pipeline: download over 12,000 PDFs, convert each page to PNG, extract 512‑D CLIP embeddings, and reduce them to 64‑D via PCA. (b) Latent‑space augmentation: concatenate the 64‑D CLIP embeddings with even objective visual‑quality metrics to form the final feature representation.}
    \label{fig:preprocessandaugment}
\end{figure}

We pass a slide thumbnail (224×224 px) through CLIP ViT‑L/14 \cite{radford2021clip}, obtaining a 512‑D embedding that jointly captures visual structure and latent semantics. To reduce redundancy and improve numerical conditioning we fit PCA on the LectureBank embeddings and retain the first 64 components, reducing from 512-D to 64-D (Figure \ref{fig:preprocessandaugment} (a)).

\subsection{Latent‑space augmentation}

\begin{figure*}[t]
    \centering
    \includegraphics[width=1\linewidth]{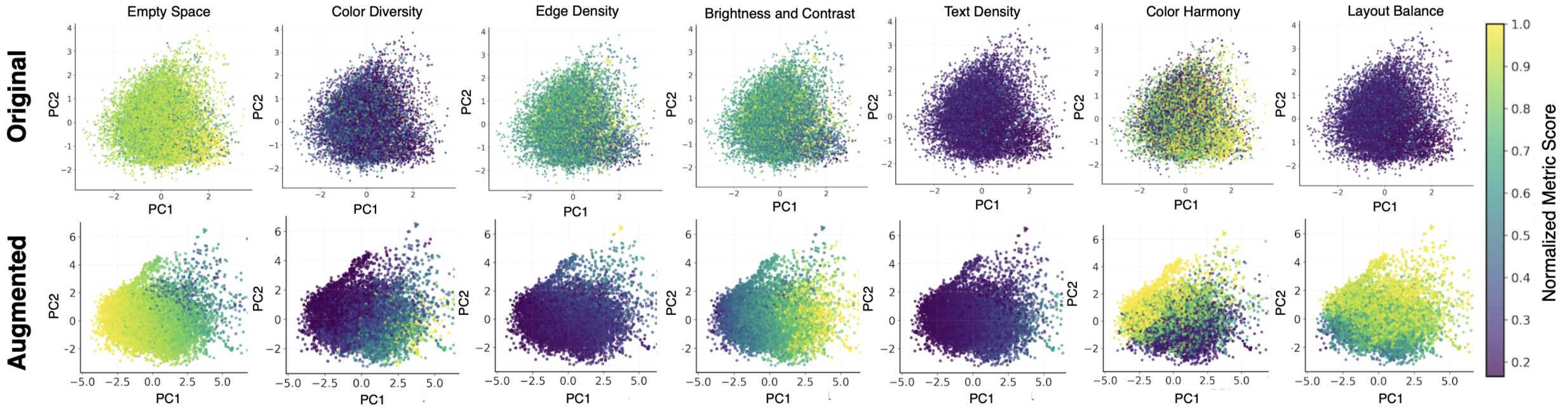}
    \caption{\textbf{Distribution of the seven slide‑quality metrics in Latent space.} Top row shows 2-D PCA projections of the original 64-D ViT latent space color-coded by each slide-quality metric; Bottom—the corresponding projections after appending the 7-D metric dimensions (augmented latent space), shown with the same color scales for direct comparison. The augmented latent space show smoother per-metric transition.}
    \label{latentfig}
\end{figure*}

We concatenate the seven scalar design‐cue metrics with the 64‑D PCA embedding to form a 71‑D slide descriptor (Figure \ref{fig:preprocessandaugment} (b)). This simple fusion yields a smoother manifold in which semantically similar slides not only cluster together in the overall latent space but also align closely along each individual metric axis, greatly facilitating anomaly detection. As shown in Figure~\ref{latentfig}, the augmented latent space both preserves the high‑level visual semantics learned by the encoder and exhibits per‑metric smoothness, as revealed by our PCA visualization.

\subsection{Unsupervised anomaly scoring}

In our approach, we treat slide quality assessment as an unsupervised outlier‑detection problem, where if a slide’s joint semantic‑and‑design signature lies far from the distribution of expert slides, we deem it low quality.

\textit{Feature vector.} For each slide we (i) encode the RGB thumbnail with CLIP ViT‑B/32, obtaining a 512‑D latent that captures layout semantics, (ii) apply PCA trained on the expert corpus and retain the first 64 components, and (iii) append seven interpretable design metrics (Whitespace, Text Density, Colorfulness, Color Harmony, Edge Density, Brightness Contrast, and Layout Balance). The resulting 71‑D vector combines high‑level content with low‑level aesthetics.

\textit{Model training} We fit an isolation forest on the slide embeddings extracted from the LectureBank corpus. Each 71‑D feature vector is z‑standardised, and the PCA projection is fit exclusively on this training set.

We adopt isolation forest defaults that have proven robust in high‑dimensional settings 
($T=200$ trees, subsample size $\psi=256$, contamination $=0.10$, \texttt{random\_state} $=42$). 
The choice $\psi=256$ tightens the theoretical bound on the expected path length $c(\psi)$, 
while 200 trees reduce variance to $<\!1\%$ \cite{liu2008isolation}. 

The resulting forest partitions the 71‑D space into hyper‑rectangles. 
Slides in dense, ‘expert’ regions require many cuts to be isolated, 
whereas poorly designed slides are separated in only a few levels, 
yielding higher anomaly scores.

\textit{Inference \& scoring.} Given a test slide $x$, we (i) project it through the same PCA and z–scaler used for training and (ii) compute the decision value:

\begin{equation}
d(x)\;=\;0.5 \;-\; 2^{-\bar{\ell}(x)\big/ c(\psi)}, 
\qquad 
\bar{\ell}(x)=\frac{1}{T}\sum_{t=1}^{T}\ell_{t}(x),
\label{eq:decision}
\end{equation}

where $\ell_{t}(x)$ is the path length of $x$ in tree $t$, $T$ is the total number of trees, and $c(\psi)$ is the expected path length for a sub‑sample of size $\psi$ \cite{liu2008isolation}. Finally, we obtain a bounded anomaly score:
\begin{equation}
a(x)=\operatorname{clip}\bigl(0.5-d(x),\,0,\,1\bigr),
\label{eq:anomaly}
\end{equation}
so that larger $a(x)$ indicates greater deviation from expert slides. Deck‑level quality is the arithmetic mean of slide‑level scores.

This single‑route pipeline meets the constraints of conference‑scale datasets: label‑free, interpretable, and computationally lightweight, yet sensitive to both semantic incongruity and design flaws.


\section{Study 1: Correlating Objective and Subjective Slide Quality}

Our anomaly‐based metric is intended to capture visual‐design quality
only. Hence, it should converge with audience ratings of slide
appearance but diverge from ratings of \textit{delivery}, which depend on prosody, gesture, and timing.  We treat delivery as a discriminant‑validity
check, while the questionnaire’s \textit{overall} item is exploratory because
it blends visual and oral factors.

\subsection{Hypotheses}
\begin{itemize}
  \item \textbf{H\textsubscript{V}} (convergent)\,: anomaly scores will correlate
        \emph{negatively} with perceived visual quality.
  \item \textbf{H\textsubscript{D}} (discriminant)\,: anomaly scores will show
        no reliable correlation with delivery ratings.
  \item \textbf{H\textsubscript{O}} (exploratory)\,: relationship with overall
        impression is reported but not preregistered.
\end{itemize}

\subsection{Subjective Measures}
Six presenters each gave a 15‑min talk on an academic paper.  
Immediately afterwards, the audience (\(N_\text{raters}=6\)) completed
4‑point Likert ratings for \textit{Visuals}, \textit{Delivery},
and \textit{Overall}.  Raters were blind to the anomaly scores.

\subsection{Scale Reliability}
Table~\ref{tab:reliability} reports internal‑consistency and agreement
statistics.  \textit{Visuals} shows strong reliability
(\(\alpha=0.85\), \(ICC=0.85\), \(W=0.50\)).
\textit{Delivery} and \textit{Overall} fall in the
``acceptable’’ range for exploratory work
(\(0.69\le ICC \le0.78\)).
Accordingly, we average each scale across raters for analysis, while
interpreting \textit{Delivery} with caution.

\begin{table}[!t]
  \caption{Reliability of the subjective rating scales}
  \label{tab:reliability}
  \centering
  \begin{threeparttable}
    \begin{tabular}{l ccc}
      \hline
      \textbf{Scale} & \textbf{Cronbach $\boldsymbol{\alpha}$} & \textbf{ICC(2,$k$)} & \textbf{Kendall $W$} \\
      \hline
      Visuals  & 0.853 & 0.848 & 0.502 \\
      Delivery & 0.698 & 0.688 & 0.355 \\
      Overall  & 0.702 & 0.782 & 0.543 \\
      \hline
    \end{tabular}
  \end{threeparttable}
  \vspace{-6pt}
\end{table}

\subsection{Analysis Procedure}
For each of the six talks we paired the mean questionnaire scores
with the slide‑deck anomaly score.  
Because Shapiro–Wilk tests indicated non‑normality
($p<0.05$), we used Spearman’s rank correlation
(\(\rho\)).  With \(n=6\) the critical magnitude for two‑tailed
\(\alpha=0.05\) is \(|\rho|\ge 0.75\).
We also report 95\% confidence intervals (Fisher’s \(z\) transform).

\begin{table}[t]
  \centering
  \caption{Spearman correlations between anomaly score and questionnaire scales}
  \label{tab:objsubcorrelation}
  \begin{tabular}{lcc}
    \toprule
    \textbf{Scale} & \(\rho\) & 95\%CI \\
    \midrule
    Visuals  & $-0.83^{*}$ & [$-0.98$, $-0.05$] \\
    Delivery & $-0.09$     & [$-0.77$, $ \phantom{-}0.71$] \\
    Overall  & $-0.49$     & [$-0.94$, $ \phantom{-}0.57$] \\
    \bottomrule
  \end{tabular}
  \begin{tablenotes}\footnotesize
    \item[*] \hspace{0.5cm}$p=.041$ (two‑tailed, \(df=4\)); CIs via Fisher \(z\).
  \end{tablenotes}
\end{table}

\subsection{Results}
\textbf{Visuals.}  A strong negative correlation supports
H\textsubscript{V}: decks nearer the expert
distribution are judged more visually appealing
(\(\rho=-0.83\), \(p=0.041\)).  
\textbf{Delivery.}  The near‑zero correlation
(\(\rho=-0.09\), \(p=0.87\)) upholds
H\textsubscript{D}\,, indicating the metric is not contaminated by
speech‑related factors.  
\textbf{Overall.}  The moderate, non‑significant trend
(\(\rho=-0.49\), \(p=0.32\)) leaves
H\textsubscript{O} unresolved.

The evidence satisfies Campbell and Fiske’s multitrait–multimethod criteria \cite{campbell1959mtmm}: high convergent and low discriminant associations. That said, the wide confidence bands reflect the small sample ($n=6$ decks $\times$ 6 raters) and should be interpreted as preliminary. Future work with larger and more diverse cohorts is required to tighten the intervals, test generalizability across slide genres, and establish reliable operating thresholds for automatic feedback systems.

\section{Study 2: Ablation study of visual representation on slide images}

\begin{figure}
     \centering
     \includegraphics[width=1\linewidth]{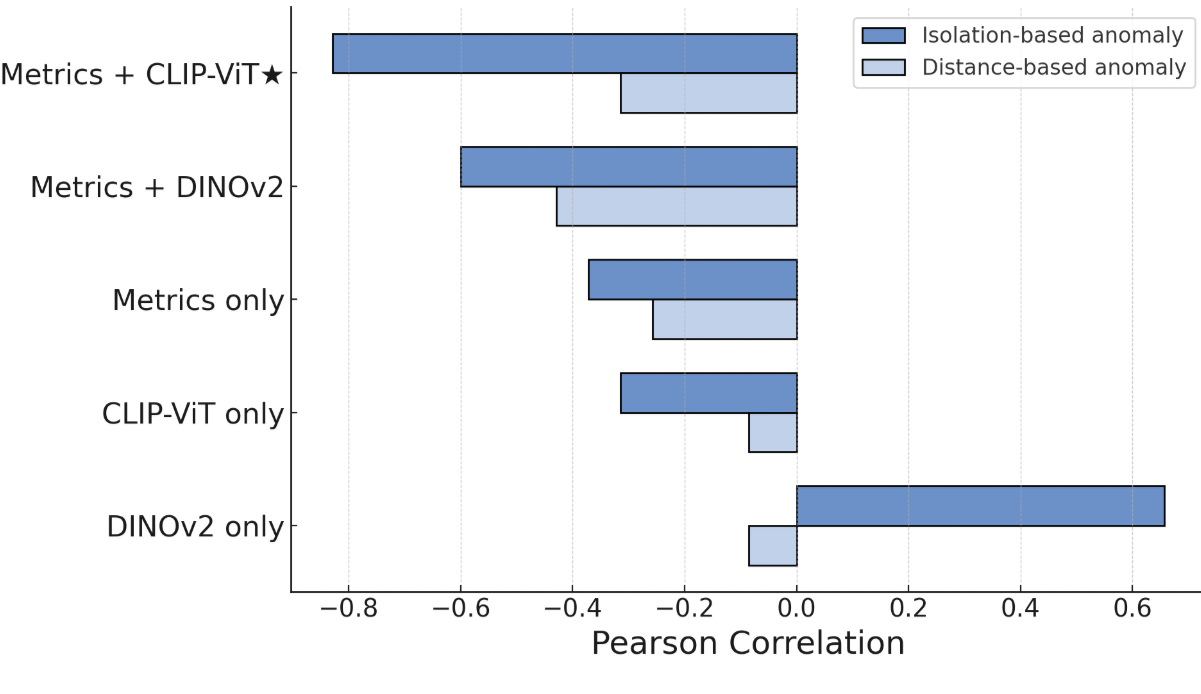}
     \caption{Ablation study of correlation strength of different visual encoder combinations and anomaly scoring methods.}
     \label{fig:ablation_bar_chart}
 \end{figure}

We compare our model with other existing pre-trained visual coders such as DINOv2\cite{oquab2024dinov2learningrobustvisual}, a large vision backbone which, unlike CLIP, is fully self-supervised and trained only on images. Same as the CLIP-based method, we also concatenate DINOv2 with the metric dimensions. 
\begin{figure}
    \centering
    \includegraphics[width=1\linewidth]{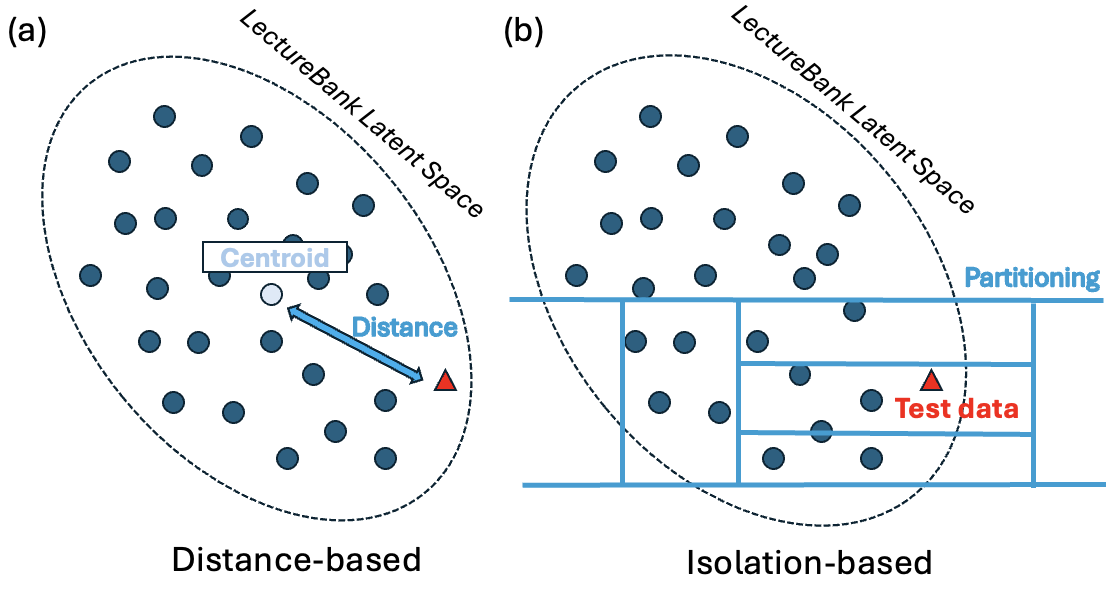}
        \caption{Comparison of two anomaly‐scoring approaches:  
        (a) \textbf{Distance‐based scoring}: compute the centroid of the reference slide distribution by averaging their feature vectors, then assign each test slide an anomaly score equal to its Euclidean distance from this centroid.  
        (b) \textbf{Isolation Forest scoring}: train an Isolation Forest on the reference set and use the average path length needed to isolate each test slide as its anomaly score, with shorter paths indicating stronger anomalies.}
    \label{fig:enter-label}
\end{figure}
Next, other methods such as calculating the euclidean distance of a data point a data cluster can help uncover anomaly behaviors \cite{6566435}. In this study, we compare the isolation forest-based method with a geometric anomaly detection by extracting the centroid of a dataset by computing the mean, and calculating the euclidean distance from a test data point. The result is summarized in Fig. \ref{fig:ablation_bar_chart}, in which the combination of metrics + CLIP-ViT and isolation forest-based anomaly scoring extracts the strongest correlation with the audience rating of the presentations. We also found that in most cases except for DINOv2 only, isolation-based anomaly detection outperforms the distance-based anomaly method in having stronger negative correlation.

\section{Study 3: Comparison with VLM methods on slide quality evaluation}
\begin{figure}[!t]
    \centering
    \includegraphics[width=1\linewidth]{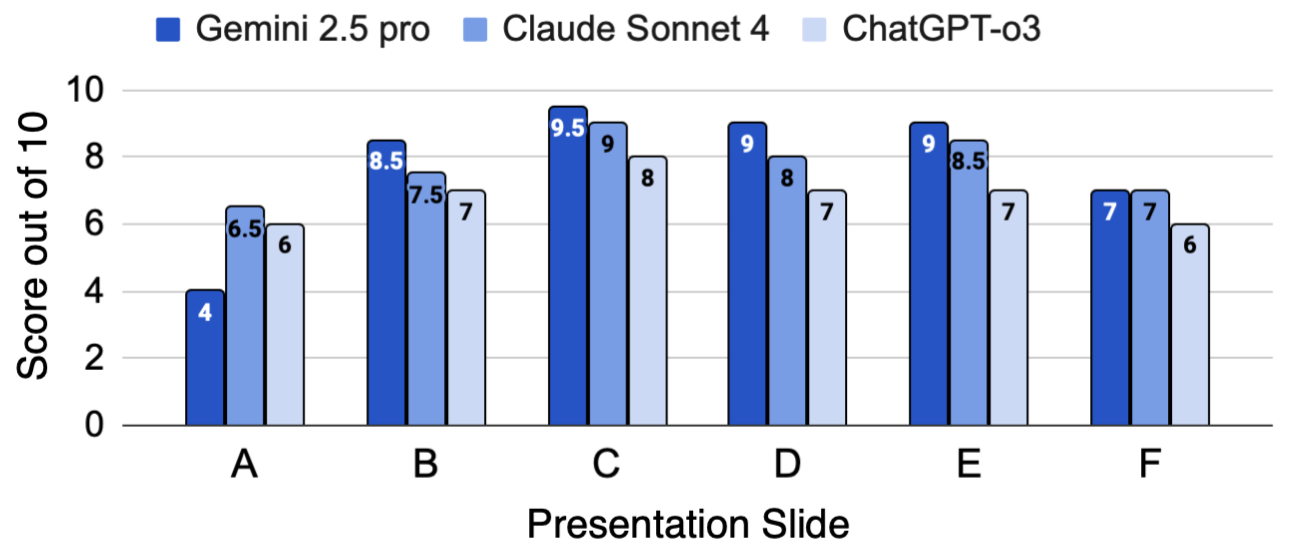}
    \caption{Scores given by popular VLM models such as Gemini 2.5 pro, Claude Sonnet 4, and ChatGPT-o3 on each of the six test presentations. A common prompt was given to each model through their web interface to obtain a 0(lowest)-to-10(highest) score.}
    \label{fig:vlm_ratings}
\end{figure}

Many recent Vision-Langauge Models (VLM) \cite{hinck2024llavagemmaacceleratingmultimodalfoundation}\cite{bai2023qwenvlversatilevisionlanguagemodel}\cite{lu2024deepseekvlrealworldvisionlanguageunderstanding} offer human-like abilities to comprehend visual information and context, which have been widely used in information retrieval and processing digital documents \cite{appalaraju2021docformerendtoendtransformerdocument}\cite{kim2022ocrfreedocumentunderstandingtransformer}.In this study, we benchmark our method against several popular vision‑language models by submitting the same test dataset—six presentation documents totaling 115 slides—to ChatGPT o4‑mini‑high, ChatGPT o3, Claude Sonnet 4, and Gemini 2.5 Pro. Using an identical prompt and no prior context, we asked each model to rate each presentation on a scale from 0 (lowest) to 10 (highest). The scores provided to each of the six presentations are reported in Fig. \ref{fig:vlm_ratings}.

We then computed the Pearson correlation between each model’s scores and the audience’s subjective evaluations (Table \ref{tab:correlation}). Our top method outperformed the VLM baselines by factors of 1.79, 2.60, and 3.23. Because our approach produces anomaly scores (where higher values indicate greater deviation), we inverted the sign to make them directly comparable to the VLM models’ ratings.

Overall, integrating objective visual quality metrics with CLIP‑ViT embeddings achieved the strongest correlation with audience evaluations. We attribute this to CLIP‑ViT’s multimodal training objective, which aligns visual features with natural‑language semantics—effectively granting the vision encoder a form of language understanding. In contrast, DINOv2 is trained purely via self‑supervised instance discrimination on images and therefore lacks that linguistic grounding, leading to its comparatively weaker alignment with subjective audience scores.

\begin{table}
  \caption{Pearson correlation ($r$) with VLM ratings}
  \label{tab:correlation}
  \centering
  \begin{threeparttable}
    \begin{tabular}{l c}
      \hline
      \textbf{Method} & \textbf{Pearson $r$} \\
      \hline
      Metrics + CLIP--ViT anomaly\tnote{*} & 0.829 \\
      ChatGPT--\textit{o}3                 & 0.463 \\
      Gemini 2.5 Pro                       & 0.319 \\
      Claude Sonnet 4                      & 0.257 \\
      \hline
    \end{tabular}
    \begin{tablenotes}[flushleft]\footnotesize
      \item[*] Absolute value reported because anomaly sign is inverted w.r.t.\ VLM ratings.
    \end{tablenotes}
  \end{threeparttable}
  \vspace{-6pt}
\end{table}

\section{Conclusion}
We have introduced a fully unsupervised slide‐quality assessment pipeline that augments seven designer‐cue metrics with CLIP‑ViT embeddings and applies Isolation Forest–based anomaly scoring to rank presentation slides. Across six test presentation slides (115 slides), our method achieved up to 3.23× stronger Pearson correlation with subjective audience ratings than off‑the‑shelf vision–language models, while demonstrating convergent validity with visual scores and discriminant validity against delivery ratings. These results confirm that combining low‐level design features with multimodal visual embeddings can approximate audience perceptions of slide quality, providing a scalable, objective tool for real‐time feedback in both educational and professional contexts.

\section{Limitation and future work}
Our evaluation is limited by a small, domain‐specific sample (six academic presentations by university students) and reliance on a lecture‐slide corpus for Isolation Forest training, which may not generalize to other slide styles or domains. We also only contrasted two pretrained backbones (CLIP‑ViT and DINOv2) and static design metrics, omitting dynamic elements such as animations or speaker–slide interplay. In future work, we will validate the pipeline on larger, more diverse slide datasets; explore additional vision–language and multimodal encoders; integrate audio, gesture, and temporal cues; and deploy interactive, in‐situ feedback systems to evaluate impact on presenter performance and audience outcomes.

\bibliographystyle{IEEEtran}
\bibliography{references}

\begin{thebibliography}{10}
\providecommand{\url}[1]{#1}
\csname url@samestyle\endcsname
\providecommand{\newblock}{\relax}
\providecommand{\bibinfo}[2]{#2}
\providecommand{\BIBentrySTDinterwordspacing}{\spaceskip=0pt\relax}
\providecommand{\BIBentryALTinterwordstretchfactor}{4}
\providecommand{\BIBentryALTinterwordspacing}{\spaceskip=\fontdimen2\font plus
\BIBentryALTinterwordstretchfactor\fontdimen3\font minus \fontdimen4\font\relax}
\providecommand{\BIBforeignlanguage}[2]{{%
\expandafter\ifx\csname l@#1\endcsname\relax
\typeout{** WARNING: IEEEtran.bst: No hyphenation pattern has been}%
\typeout{** loaded for the language `#1'. Using the pattern for}%
\typeout{** the default language instead.}%
\else
\language=\csname l@#1\endcsname
\fi
#2}}
\providecommand{\BIBdecl}{\relax}
\BIBdecl

\bibitem{structureness2023}
\BIBentryALTinterwordspacing
T.~Suwartono, ``Presentation content structuredness: How does it affect the audience?'' \emph{ResearchGate}, 2023. [Online]. Available: \url{https://www.researchgate.net/publication/370123456}
\BIBentrySTDinterwordspacing

\bibitem{garner2012assertion}
\BIBentryALTinterwordspacing
J.~K. Garner and M.~Alley, ``How the design of presentation slides affects audience comprehension: A case for the assertion-evidence approach,'' \emph{International Journal of Engineering Education}, vol.~28, no.~1, pp. 156--164, 2012. [Online]. Available: \url{https://www.researchgate.net/publication/285592570}
\BIBentrySTDinterwordspacing

\bibitem{blaseg2025visualpitch}
D.~Blaseg and S.~Mayer, ``Judging a pitch by its cover? evidence on the roles of visual fluency and substantive quality in startup pitch decks,'' in \emph{Proceedings of the BSE Summer Forum Workshop on Entrepreneurship}, 2025, field and online experiments with nearly 34,000 participants.

\bibitem{nielsen1994heuristics}
\BIBentryALTinterwordspacing
J.~Nielsen, ``10 usability heuristics for user interface design,'' 1994, accessed: 2025-06-30. [Online]. Available: \url{https://www.nngroup.com/articles/ten-usability-heuristics/}
\BIBentrySTDinterwordspacing

\bibitem{mayer2005cognitive}
R.~E. Mayer, ``Cognitive theory of multimedia learning,'' in \emph{The Cambridge Handbook of Multimedia Learning}, R.~E. Mayer, Ed.\hskip 1em plus 0.5em minus 0.4em\relax Cambridge University Press, 2005, pp. 31--48.

\bibitem{naegle2021slides}
K.~M. Naegle, ``Ten simple rules for effective presentation slides,'' \emph{PLOS Computational Biology}, vol.~17, no.~12, p. e1009554, 2021.

\bibitem{zhang2022color}
L.~Zhang, W.~Sun \emph{et~al.}, ``Perceptual modeling of color harmony: A 16-dimensional feature model for automatic evaluation,'' \emph{Frontiers in Psychology}, vol.~13, p. 876543, 2022.

\bibitem{zhou2022colorharmony}
J.~Zhou, F.~Liu, and S.~Li, ``Attribute analysis and modeling of color harmony based on multi-dimensional physical features,'' \emph{Frontiers in Psychology}, vol.~13, p. 998650, 2022.

\bibitem{rosenholtz2007measuring}
R.~Rosenholtz, Y.~Li, and L.~Nakano, ``Measuring visual clutter,'' \emph{Journal of Vision}, vol.~7, no.~2, p.~17, 2007.

\bibitem{ngo2003modelling}
D.~C. Ngo, L.~S. Teo, and J.~G. Byrne, ``Modelling interface aesthetics,'' \emph{Information Sciences}, vol. 152, pp. 25--46, 2003.

\bibitem{kim2014classification}
S.~Kim, W.~Jung, K.~Han, J.-G. Lee, and M.~Y. Yi, ``Quality-based automatic classification for presentation slides,'' in \emph{Proceedings of the 36th European Conference on Information Retrieval}.\hskip 1em plus 0.5em minus 0.4em\relax Springer, 2014, pp. 638--643.

\bibitem{multimodal2017}
A.~Hanani, M.~Al-Amleh, W.~Bazbus, and S.~Salameh, ``Automatic estimation of presentation skills using speech, slides and gestures,'' in \emph{Speech and Computer}, A.~Karpov, R.~Potapova, and I.~Mporas, Eds.\hskip 1em plus 0.5em minus 0.4em\relax Cham: Springer International Publishing, 2017, pp. 182--191.

\bibitem{leceval2024}
\BIBentryALTinterwordspacing
X.~Liu, Z.~Zhao, W.~Ni, Y.~Wang, Y.~Wu, Y.~He, W.~Sun, and L.~Zhang, ``Leceval: Interpretable slide evaluation for lecture scenario with multimodal learning,'' \emph{arXiv preprint arXiv:2505.02078}, 2024. [Online]. Available: \url{https://arxiv.org/abs/2505.02078}
\BIBentrySTDinterwordspacing

\bibitem{pptagent2024}
\BIBentryALTinterwordspacing
S.~Gao, S.~Qiao, Z.~Wang, Z.~Wang, J.~Zhang, X.~V. Lin, C.~Liu, W.~Yu, T.~Zhao, P.~Liang, and T.~Ma, ``Ppteval: Benchmarking large language models on slide deck evaluation,'' \emph{arXiv preprint arXiv:2501.03936}, 2024. [Online]. Available: \url{https://arxiv.org/abs/2501.03936}
\BIBentrySTDinterwordspacing

\bibitem{li2018lecturebank}
\BIBentryALTinterwordspacing
Irene Li, Alexander R. Fabbri, Robert R. Tung, and Dragomir R. Radev, ``What should i learn first: Introducing lecturebank for nlp education and prerequisite chain learning,'' \emph{CoRR}, vol. abs/1811.12181, 2018. [Online]. Available: \url{http://arxiv.org/abs/1811.12181}
\BIBentrySTDinterwordspacing

\bibitem{radford2021clip}
\BIBentryALTinterwordspacing
A.~Radford, J.~W. Kim, C.~Hallacy, A.~Ramesh, G.~Goh, S.~Agarwal, G.~Sastry, A.~Askell, P.~Mishkin, J.~Clark, G.~Krueger, and I.~Sutskever, ``Learning transferable visual models from natural language supervision,'' in \emph{Proceedings of the 38th International Conference on Machine Learning (ICML)}, ser. Proceedings of Machine Learning Research, vol. 139.\hskip 1em plus 0.5em minus 0.4em\relax PMLR, 2021, pp. 8748--8763. [Online]. Available: \url{http://proceedings.mlr.press/v139/radford21a.html}
\BIBentrySTDinterwordspacing

\bibitem{liu2008isolation}
F.~Liu, K.~Ting, and Z.~Zhou, ``Isolation forest,'' in \emph{Proceedings of the 2008 IEEE International Conference on Data Mining}.\hskip 1em plus 0.5em minus 0.4em\relax IEEE, 2008, pp. 413--422.

\bibitem{campbell1959mtmm}
D.~T. Campbell and D.~W. Fiske, ``Convergent and discriminant validation by the multitrait–multimethod matrix,'' \emph{Psychological Bulletin}, vol.~56, no.~2, pp. 81--105, 1959.

\bibitem{oquab2024dinov2learningrobustvisual}
\BIBentryALTinterwordspacing
M.~Oquab, T.~Darcet, T.~Moutakanni, H.~Vo, M.~Szafraniec, V.~Khalidov, P.~Fernandez, D.~Haziza, F.~Massa, A.~El-Nouby, M.~Assran, N.~Ballas, W.~Galuba, R.~Howes, P.-Y. Huang, S.-W. Li, I.~Misra, M.~Rabbat, V.~Sharma, G.~Synnaeve, H.~Xu, H.~Jegou, J.~Mairal, P.~Labatut, A.~Joulin, and P.~Bojanowski, ``Dinov2: Learning robust visual features without supervision,'' 2024. [Online]. Available: \url{https://arxiv.org/abs/2304.07193}
\BIBentrySTDinterwordspacing

\bibitem{6566435}
M.~Ahmed and A.~N. Mahmood, ``A novel approach for outlier detection and clustering improvement,'' in \emph{2013 IEEE 8th Conference on Industrial Electronics and Applications (ICIEA)}, 2013, pp. 577--582.

\bibitem{hinck2024llavagemmaacceleratingmultimodalfoundation}
\BIBentryALTinterwordspacing
M.~Hinck, M.~L. Olson, D.~Cobbley, S.-Y. Tseng, and V.~Lal, ``Llava-gemma: Accelerating multimodal foundation models with a compact language model,'' 2024. [Online]. Available: \url{https://arxiv.org/abs/2404.01331}
\BIBentrySTDinterwordspacing

\bibitem{bai2023qwenvlversatilevisionlanguagemodel}
\BIBentryALTinterwordspacing
J.~Bai, S.~Bai, S.~Yang, S.~Wang, S.~Tan, P.~Wang, J.~Lin, C.~Zhou, and J.~Zhou, ``Qwen-vl: A versatile vision-language model for understanding, localization, text reading, and beyond,'' 2023. [Online]. Available: \url{https://arxiv.org/abs/2308.12966}
\BIBentrySTDinterwordspacing

\bibitem{lu2024deepseekvlrealworldvisionlanguageunderstanding}
\BIBentryALTinterwordspacing
H.~Lu, W.~Liu, B.~Zhang, B.~Wang, K.~Dong, B.~Liu, J.~Sun, T.~Ren, Z.~Li, H.~Yang, Y.~Sun, C.~Deng, H.~Xu, Z.~Xie, and C.~Ruan, ``Deepseek-vl: Towards real-world vision-language understanding,'' 2024. [Online]. Available: \url{https://arxiv.org/abs/2403.05525}
\BIBentrySTDinterwordspacing

\bibitem{appalaraju2021docformerendtoendtransformerdocument}
\BIBentryALTinterwordspacing
S.~Appalaraju, B.~Jasani, B.~U. Kota, Y.~Xie, and R.~Manmatha, ``Docformer: End-to-end transformer for document understanding,'' 2021. [Online]. Available: \url{https://arxiv.org/abs/2106.11539}
\BIBentrySTDinterwordspacing

\bibitem{kim2022ocrfreedocumentunderstandingtransformer}
\BIBentryALTinterwordspacing
G.~Kim, T.~Hong, M.~Yim, J.~Nam, J.~Park, J.~Yim, W.~Hwang, S.~Yun, D.~Han, and S.~Park, ``Ocr-free document understanding transformer,'' 2022. [Online]. Available: \url{https://arxiv.org/abs/2111.15664}
\BIBentrySTDinterwordspacing

\end{thebibliography}

\end{document}